\documentclass[final,5p,times,twocolumn,authoryear,preprint]{elsarticle}

\usepackage[english]{babel}
\usepackage[T1]{fontenc}
\usepackage[utf8]{inputenc}

\usepackage{graphicx}
\usepackage{amssymb}

\usepackage{algorithm}
\usepackage{algorithmic}
\usepackage{multirow}
\biboptions{comma}

\begin{document}

\begin{frontmatter}

\title{BIN-CT: Urban Waste Collection based on Predicting the Container Fill Level}

\author{Javier Ferrer\corref{cor1}}
\ead{ferrer@lcc.uma.es}
\author{Enrique Alba}
\ead{eat@lcc.uma.es}

\cortext[cor1]{Corresponding author}

\address{Dept. Lenguajes y Ciencias de la Computaci\'on, University of Malaga, Spain}

\begin{abstract}
The fast demographic growth, together with the population concentration in cities and the increasing amount of daily waste, are factors that are pushing to the limit the ability of waste assimilation by Nature. Therefore, we need technological means to optimally manage of the waste collection process, which represents 70\% of the operational cost in waste treatment. In this article, we present a free intelligent software system called BIN-CT (BIN for the CiTy), based on computational learning algorithms, which plans the best routes for waste collection supported by past (historical) and future (predictions) data.

\noindent The objective of the system is to reduction the cost of the waste collection service minimizing the distance traveled by a truck to collect the waste from a container, thereby reducing the fuel consumption. At the same time the quality of service for the citizen is increased, avoiding the annoying overflows of containers thanks to the accurate fill-level predictions given by BIN-CT. In this article we show the features of our software system, illustrating its operation with a real case study of a Spanish city. We conclude that the use of BIN-CT avoids unnecessary trips to containers, reduces the distance traveled to collect a container by 20\%, and generates routes 33.2\% shorter than the routes used by the company. Therefore it enables a considerable reduction of total costs and harmful emissions thrown up into the atmosphere.
\end{abstract}


\end{frontmatter}

\section{Introduction}

Waste generation is globally growing at an unstoppable rate, in contrast, the assimilation capacity of our planet is decreasing, turning waste treatment into a complex challenge we need to address as a modern society \citep{Al-Salem2009}. In addition, the linear structure of our economy (extract, manufacture, use and disposal) has reached its limits; the Earth is exhausted, natural resources are drained. The unsustainable development of most nations has created a problem due to the growing generation of waste that must be solved. Despite the social and governmental commitment, there are hardly any technological means to optimise the management of the waste collection process. The largest resources are needed to collect the municipal waste, which is mainly created by household, industrial, and commercial activity.

Traditionally solid waste collection was carried out without previously analyzing the demand or the routes of the vehicles and the decisions about the routes were made by the drivers, although the solutions were far from optimal. Solid waste collection is usually undertaken based on static plans, with a pre-determined number of trips per week, designed manually in most cases. This approach has severe limitations due to the number of constraints to consider when finding an optimal solution. Moreover, waste collection planning problems are influenced by stochastic waste generation, traffic conditions, and many constraints, which are unmanageable by a person, so the use of an automatic tool to solve these types of problems is mandatory. 


In this article we propose an intelligent software system, called BIN-CT, for the management of solid waste collection in an urban area. The system is doubly intelligent, since it integrates algorithms to predict the fill level of waste containers, plus the subsequent optimal generation of routes for collection trucks. In other words, our software system solves the two major problems that a waste collection service face: 1) what containers should be collected and 2) in what order should the containers be visited to minimize the cost.

In order to solve the first question, we will use computational machine learning techniques to estimate the fill level of each container. Supported by historical data (past data), the system generates predictions (future data) to decide when a container should be collected. This prediction could be performed long term or short term, considering seasonality, weekends, and holidays. We want to highlight that, as far as we know, the fill-level prediction of all containers individually (fine grain) is a feature, which until now, has not been implemented in any tool for waste collection management. Additionally, our software system is able to interact with an Internet of Things (IoT) system to obtain the actual filling of the containers equipped with volumetric sensors. Moreover, the sensors' information will be essential in the near future to validate the historical data (obtained by truck drivers). \\

Route generation (of waste collection in our case) is a well-known combinatorial problem called the Vehicle Routing Problem (VRP)~\citep{Dantzig1959}. However the use of historical data, predictions and the intelligent decisions about the inclusion/exclusion of containers mean that it is not a simple VRP. The solutions to our problem are usually constrained by many factors. Some of them are the number of available trucks, the trucks' capacity, the amount of waste to collect and the characteristics of the street where the container is placed (too narrow for a big truck), among others. The planning of waste collection routes has a wide number of variants and constraints which makes it unmanageable for one person alone. In contrast, there is great room for improvement using automatic algorithms to deal with this problem. The improvement results in lower cost services, a reduction of harmful emissions and a better service for citizens, in quality and costs. 

In addition, we must take into account that the current city system itself is under discussion. Many local authorities have been forced to examine its cost-effectiveness and environmental impact since we have a system based on static collection frequencies (once a week, twice a week, daily…) that sometimes results in unnecessary trips to semi-empty containers. The pollution generated by these trips could be more dangerous for the environment than the benefits of the collection. This is especially critical in the case of selective collection (plastic, paper, glass…), where the waste volume is smaller than in the case of organic waste. So, when you are dealing with recyclable waste, the planning of optimal collection routes is even more influential. Presently, the recyclable waste collection process represents 70\% of the operational cost of waste treatment~\citep{Teixeira2004}. The main objective of the work presented here is the reduction of the collection process cost and is essentially dependent on the distance travelled by the collection trucks. 

\begin{figure}         \centering   \vspace{0.2cm}   \includegraphics[width=5cm]{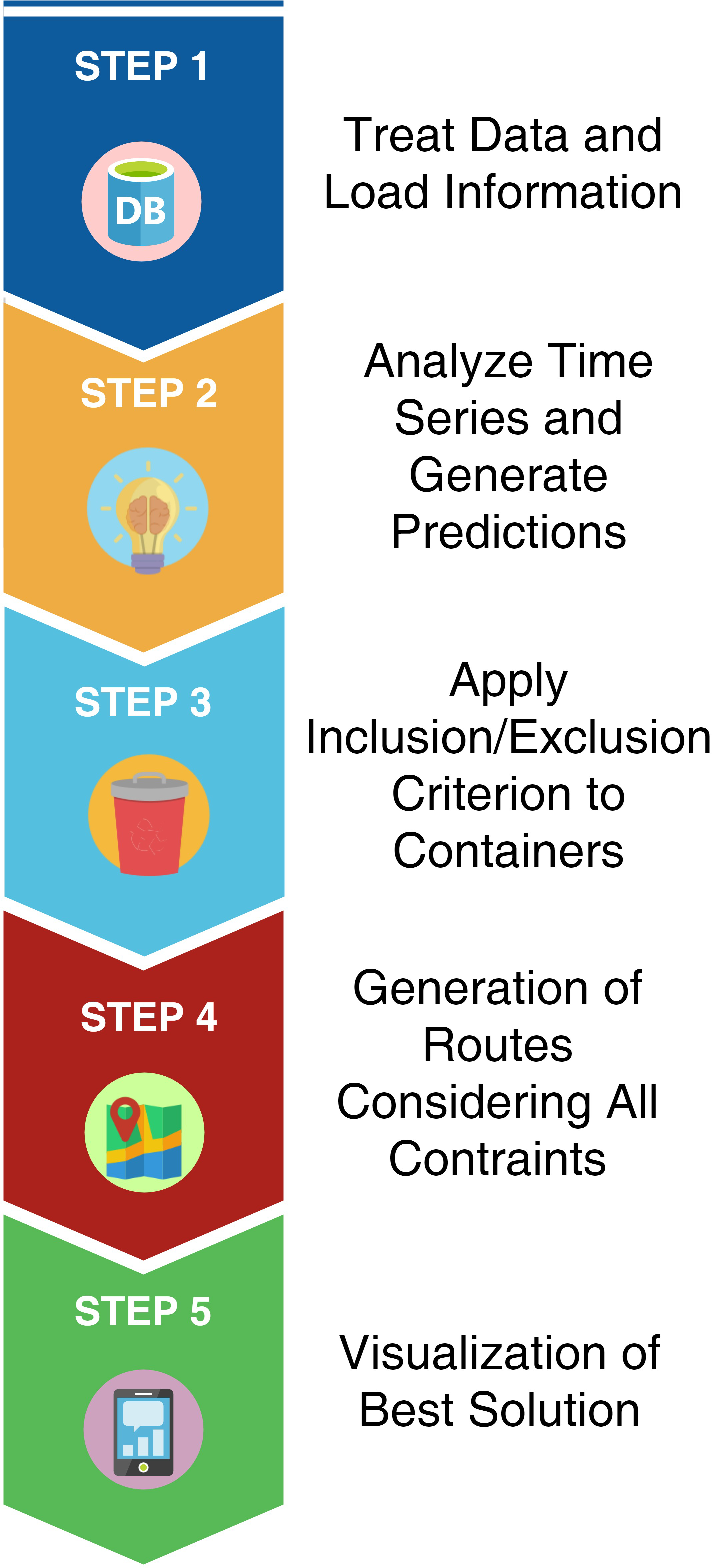}         \caption{Methodology followed to solve the waste collection problem.}
\label{fig:scheme}     \end{figure}

In Figure~\ref{fig:scheme} we show the steps followed to solve the problem we face in this article. The main phases are as follows: First, all the data needed to solve the problem is treated and stored in a database. In this way, the BIN-CT system is able to load the information from the database instead of using several large files. In the second phase, the software system uses the historical fill-level data of each container and generates the fill-level predictions. In the third step, the software system decides which containers are going to be included in the next routes. The decision is taken according to a given criterion based on the predictions and estimated fill levels. In the fourth phase the route algorithm computes optimal solutions, i.e., the complete routes that the trucks must follow to collect the containers selected in the previous step. Finally, the best solution is shown on a navigable map in a web browser. 

The main contributions of this article are as follows: \begin{itemize} \item We propose an intelligent system called BIN-CT for generating optimal waste collection routes.
\item We compare three machine-learning algorithms for the forecasting of filling levels of each container (fine grain) from historical data.\\
\item We use a new real case study of a city with 217 containers to illustrate the behavior of the proposed system.
\item 
We compare the results obtained by BIN-CT to the results of a real company in a real-world case study.

\end{itemize}
The remainder of the article is as follows. In Section~\ref{related} we describe work related to waste collection optimization, especially the different variants studied in the literature. Section~\ref{problem} describes the waste collection problem we address in this article. Section~\ref{binct} explains our proposal, the BIN-CT software system. The four modules that comprise our software system are described and we outline the most important features of the system. Our experimental setup is presented in Section~\ref{experiments}, then we describe the real-world case study, and analyze the results of two different experiments. Finally, we present our conclusions and future work in Section~\ref{conclusions}.

\section{Related Work}
\label{related}

Waste collection is a process with countless variants and constraints which have led to a multitude of studies in recent years due to its importance. The approaches in the literature could be classified, among other ways, according to the waste type that is treated. Some of them are: \textit{residential waste} commonly known as garbage~\citep{GCD11}, \textit{industrial waste} where customers/containers are more dispersed and the amount of waste is higher~\citep{Rabbani2018}, \textit{recyclable waste}~\citep{Dat2012} increasingly important for our society, where the collection frequency is lower than organic waste and \textit{hazardous waste} where the probability of harm and costs are both minimized~\citep{Rabbani2018}.

In municipal solid waste collection~\citep{Belien2014}, the authorities need global studies to quantify the waste generated in a period of time to be able to manage it. For example, waste generation forecasting for the inhabitants of Xiamen City (China) was studied by~\citep{Xu2013}. The main difference with our approach is the granularity of the object under study. They predict the amount of waste produced in the whole city, in contrast, we generate predictions for every single container in a city. 

This supposes a considerable increase in the complexity of the problem that is to be solved, because it is necessary to consider multiple aspects such as the location, the customs of the citizens, the population density of the area, etc. Along the same research lines, the impact of the intervention of local authorities on waste collection has also been studied \citep{Cole2014}, with this being relevant in the medium-long term.

Regarding the location where the collection takes place, there exist multiple variants of the problem. There are \textit{communal collections} where the local authority identifies a place shared by the community~\citep{Tung2000}, in most cases a local waste facility for recycling. On the other side, we found the \textit{kerbside collection}~\citep{Sniezek2006} where the household waste is collected from individual small containers located near each house. The intermediate case studied here is the collection of containers which serve several streets and blocks of flats~\citep{Sakti2018}. 

In medium/large cities, the number of containers is in the range of hundreds or even thousands, therefore the use of advanced computational methods are required to find an optimal solution. In the past, exact methods were used to solve this problem like \textit{Branch and Bound}, based on \textit{mathematical programming}~\citep{Arribas2010}. However the execution time tends to increase exponentially with the number of containers, so the use of hybrid methods like matheuristics~\cite{Gruler2017} and bioinspired methods is recommended. Although some metaheuristics have already been explored in the literature such as \textit{Ant Colony optimization}, \textit{Genetic Algorithms}~\citep{Buenrostro-Delgado2015}, \textit{Greedy Randomized Adaptive Search Procedure}~\cite{Gruler2017-2}, or \textit{Variable Neighborhood Search}~\cite{Gruler2017-3}. There is still room for improvement when one is dealing with the waste collection problem at the fine granularity we address it in our approach, that is, analyzing each of every single container.

The waste collection process implies a large quantity of stakeholders like authorities, citizens, the company in charge of the collection service and the company workers. Depending on the stakeholder, the objectives to optimize are different. Some examples of optimization goals are the \textit{number of vehicles} for the service~\citep{Ombuki-Berman2007}, the service \textit{total cost}~\citep{Arribas2010}, the \textit{environmental impact}~\citep{Tavares2009}, the \textit{required staff} for the service~\citep{Baudach}, the collection \textit{route length}~\citep{Ustundag2008}, and the \textit{total time} of the waste collection~\citep{Arribas2010}. Our approach considers several objectives at a time, finding solutions which minimize the number of vehicles needed, the total cost, the route length, and the environmental impact. 
 
\section{Problem Description}
\label{problem}

Waste collection management is a global problem that affects most cities around the world. In fact, this problem falls within the initiative of Horizon 2020 and Smart City for innovation and research, promoted by the European Union. The reduction of energy consumption and the responsible use of resources has become key to combat the rigors of the economic crisis. Specifically, one of the objectives of the Smart City initiative is to reduce the emission of greenhouse gases, use sustainable resources, and efficiently manage energy sources. We are therefore dealing with a problem of general interest in this article. We focus on some of the aspects discussed in Section~\ref{related}. Specifically, we study recyclable waste, we use fine grain predictions for every single container, we manage hundreds of containers, and we use an evolutionary algorithm as resolution technique.

The problem consists of planning how the waste collection will take place in a determined area. The objective is to define optimal collection routes for the available trucks whose capacity cannot be exceeded. An optimal collection route is an ordered list of collection places to visit that minimizes the total distance traveled by all the trucks involved.

Since the number of containers is great for all of them to be collected every day and it is not a cost-effective strategy, the containers must be visited at a more appropriate time (ideally when their filling level is close to 100\%). In addition, routes are constrained by the trucks' capacities, that may differ. Moreover, there exist constraints which affect specific containers that must be collected by a small truck (with less capacity) due to space restrictions. Route duration is determined by the travel time between containers and the time to empty them.

Generally, there are some difficulties in estimating the distances and duration of trips between containers due to traffic conditions. Most approaches~\citep{Nuortio2006,Teixeira2004} only use the distance as a criterion for generating the cost matrix. They calculate the shortest paths between pairs of waste containers using the Dijkstra algorithm \citep{Dijkstra1959}. In contrast, we have opted to calculate the estimated distance and duration between pairs of containers using the Google Maps Api v3. By doing so, we obtain a more accurate value for travel distance and duration taking into account traffic regulations and conditions. 

The quantity deposited daily in each container is unknown, it depends on multiple factors such as the number of citizens sharing a container, the seasonality and lifestyle. In other papers, the authors assume the quantity deposited in the containers is constant every day, however, BIN-CT treats each container as a different entity that can change daily. As the previous fill levels of the containers are available, the software system forecasts future fill levels based on historical data. This information will help determine which containers must be collected in the next shift. The predictions generation is really complex, since we need to model the behavior of the population that deposits its waste in a specific container.

Overall, the complexity of this problem can be derived from the number of restrictions that can be applied to the model, so that the more restrictions applied, the more realistic the solution found and the larger its computational complexity. The constraints we take into account are highly diverse. With respect to the trucks, we can restrict the capacity of each vehicle individually, the number of them and the ability to collect containers in narrow streets. The quality of service defined by the waste management company (imposed by the local authorities in the contract with the company) could be seen as a soft constraint. In this paper the company establishes that all the containers with a filling estimation greater than 80\% must be collected. 
In summary, we enumerate the constraints considered here: 
\begin{itemize} 
\item Containers: \begin{itemize} \item Variable number of containers \item Different demands \item Different locations \item Custom unloading time \item Only specific trucks can unload a particular container \end{itemize} \item Vehicles: \begin{itemize} \item Variable number of vehicles \item Different loading capacities \item Ability to unload specific containers (narrow streets) \item Custom cost based on distance traveled \end{itemize} \end{itemize}

The vehicle routing problem is well known to be NP-hard and the waste collection problem described here is basically a VRP with several constraints, so it is even harder to solve. Furthermore, a realistic instance involving 50 or more containers and several vehicles, means thousands of decision variables and constraints~\cite{Varone2014}. Therefore the use of heuristics to obtain quasi-optimal solutions in a moderate time is recommended. In this paper, we use evolutionary algorithms for this purpose.

\subsection{Problem Formulation}

The underlying VRP in the waste collection problem is a combinatorial problem whose ground set is the edges of a graph ${G(C,A)}$. The notation used for this problem formulation is as follows:

\begin{itemize}
\item ${C = \left\lbrace c_{0}, c_{1}, …, c_{n}, c_{n+1} \right\rbrace}$ is a container set, where a depot (0 denotes the depot) is located at ${c_0}$ and ${c_{n+1}}$, and ${C’ = C \backslash \left\lbrace c_{0},c_{n+1} \right\rbrace}$ is the set of ${n}$ containers.

\item ${A = \left\lbrace(c_{i},c_{j}) | c_{i},c_{j} \in C; i \neq j \right\rbrace}$ is a directed arc set.

\item ${D}$ is a matrix of non-negative distances (meters) ${d_{ij}}$ between containers ${c_{i}}$ and ${c_{j}}$.
\item $T$ is matrix of non-negative duration (seconds) ${t_{ij}}$ between containers ${c_{i}}$ and ${c_{j}}$.
\item ${R_{i}}$ is the route for vehicle ${i}$.
\item ${V_i}$ is the vehicle $i$ which has a maximum capacity $v_i$.
\item ${v}$ is the number of vehicles. One route is assigned to each vehicle.
\item $s$ is a binary vector, which indicates if $V_i$ has the size to access the location of certain containers(e.g. those located in the city center).
\item $w$ is a binary vector, which indicates if the container $c_i$ must be collected by a vehicle $V_i$ | $s_i=1$ .
\end{itemize}


In a real city the distance between a container $i$ and $j$ could be different due to traffic regulations, therefore when $\exists i,j\;D_{ij}\neq D_{ji}$ the problem is said to be asymmetric.

Each vertex ${c_{i}}$ in ${C’}$ has an associated quantity ${q_{i}}$ of waste to be loaded by a vehicle. The VRP thus consists in determining a set of ${v}$ vehicle routes of minimal total cost, starting and ending at a depot, such that the maximum waste is collected and a container is visited at most once. Note that each vehicle has an associated maximum load capacity $v_i$ which can not be exceeded, i.e., the total quantity of waste of all containers serviced on a route ${R_{i}}$ must not exceed the vehicle capacity ${v_i}$. 

A feasible solution is composed of a set of routes ${R_{1},\dots,R_i, \dots, R_{v}}$ where $R_i$ is the route for the vehicle $i$ and a permutation ${\sigma_{i}}$ of ${R_{i} \bigcup {0}}$ (0 denotes the depot) specifying the order of the containers on route ${i}$. In addition, each route has to fulfill the following constraint related to the size of the containers: if $\exists\;c_j \in R_{i}$ such that $w_j = 1$ (e.g. the container is in the city center), then $s_i = 1$ (the vehicle has the size to reach those containers). 

Additionally, we consider a service time ${\delta_{i}}$ (time needed to load the container), required by a vehicle to load the quantity ${q_{i}}$ at container ${c_{i}}$. In this context, the time duration $\gamma_{i}$ to collect ${c_{j}}$ is the service time ${\delta_{i}}$ plus the travel time ${t_{ji}}$. Note that in this approach $\gamma_{i}$ (costs associated with the travel time and service time) are used to report an estimation of the duration of the routes.

The cost of a given route (${R_{i} = \left\lbrace c_{0}, c_{1}, …, c_{v+1} \right\rbrace}$), where ${c_{i} \in C}$ and ${c_{0} = v_{c+1} = 0}$ (0 denotes the depot), is given by 

\begin{equation}
C(R_{i}) = \sum_{i=0}^{v} D(c_i,c_{i+1})    
\end{equation}

Finally, the fitness for the problem solution ${S}$ is given by: 
\begin{equation}
{Fitness_{S} = \sum_{i=1}^{v} C(R_{i})} + p*nac
\end{equation}
where $p$ is a penalty value and $nac$ is the number of unassigned containers (due to vehicles' maximum capacity constraint). In this paper we set $p=500$, a high enough value for this instance.

\section{Our Proposal: BIN-CT Predictive System}
\label{binct}

 \begin{figure*}[]
        \centering
        \includegraphics[width=15cm]{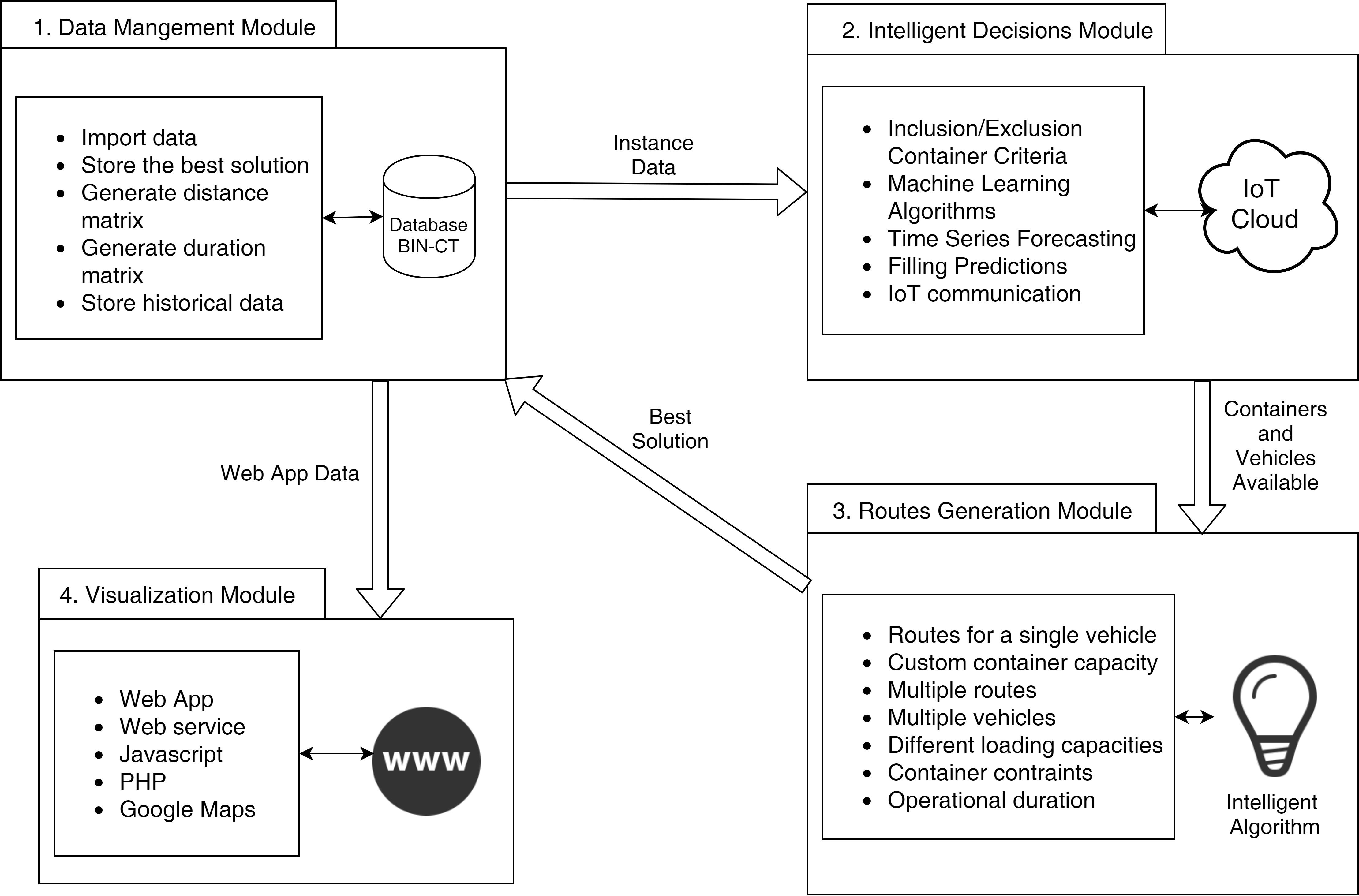}
        \caption{BIN-CT module diagram.}
        \label{fig:modules}
\end{figure*}

The BIN-CT predictive system has been designed to improve municipal waste collection planning. Our software system is a combination of two main algorithms, one for predicting the filling levels of the containers and one for computing the best routes to visit them. These algorithms are encoded into a software package that offers a Graphical User Interface (GUI) so that a manager in the recycling company can make informed decisions and take action, like giving daily routes to truck drivers to follow or pursuing new offline studies like analyzing the evolution in time of the filling level of the containers.

To deal with the complexity of the global problem, we address it in four sequential phases. In the first phase, the travel distances and duration for all pairs of containers are computed to give more realism to the solution. Then, in the second phase, the fill levels are estimated and the containers which are not going to be part of the routes for the next service are discarded. In the third phase the routes are generated, minimizing the total cost, and in the last phase we show the best solution in a web browser. Therefore, we divided the tool into four different modules which can be seen in Figure~\ref{fig:modules}.

In this section we detail the main features of our software, as well as the solution quality it achieves when solving an instance of the solid waste collection problem described here. The BIN-CT predictive system comprises four software modules: A Data Management module, Intelligent Decision module, Routes Generation module and Visualization module. 

\subsection{Data Management Module}

\begin{figure}[h!]
        \centering
        \includegraphics[width=9cm]{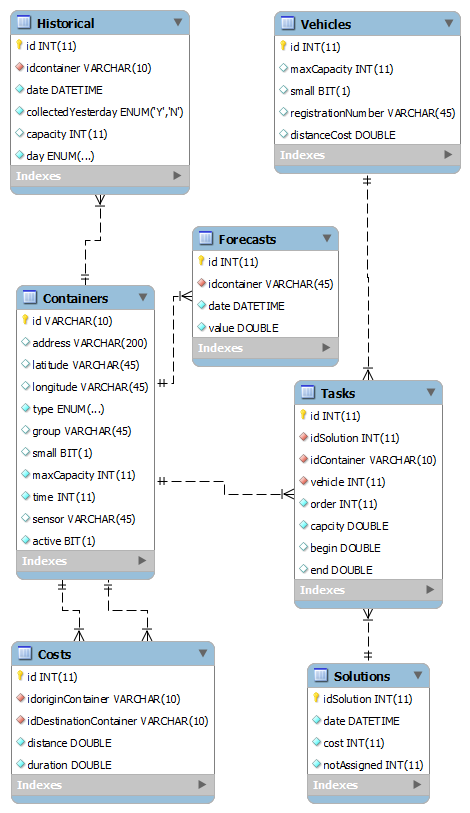}
        \caption{BIN-CT database diagram. We store the information in a clear and structured way in seven tables.}
        \label{fig:DB}
    \end{figure}

All information for the execution of the software that comes from the waste management company, which undertakes the collection, can be loaded from a database. The use of a database is desirable due to the amount of information needed to solve the problem. The main entities (input and output data) used in our system are shown in Figure~\ref{fig:DB}.  Additionally, we list the input data needed to run BIN-CT effectively in Table~\ref{tab-data}.

\begin{table}[]
\caption{Input data for the BIN-CT system.}
\label{tab-data}
\begin{tabular}{ll}
\hline
Entity & Parameters \\ \hline
Containers & \begin{tabular}[c]{@{}l@{}}Identifier\\ Location (latitude, longitude)\\ Max Capacity\\ Volumetric Sensor (Yes or No)\\ Unload Time\\ Address\\ Group\\ Small (Yes or No)\\ Sensor Identifier\end{tabular} \\ \hline
Historical Data & \begin{tabular}[c]{@{}l@{}}Container Identifier\\ Date\\ Collected Yesterday (Yes or No)\\ Day\\ Waste Collected\end{tabular} \\ \hline
Vehicles & \begin{tabular}[c]{@{}l@{}}Max Capacity\\ Small (Yes or No)\\ Registration Number\\ Costs\end{tabular} \\ \hline
\end{tabular}
\end{table}

The main entity is \textit{Containers}. We store the identification of the container, information about the location such as coordinates and addresses, the maximum capacity of the container, how long it takes to collect the container, the type of waste (glass, paper,...), the type of the truck needed to collect the container (e.g., small truck in narrow central streets) and whether it has a sensor or not to measure the fill-level.

The container is related to the historical filling data entity where we store the filling data of each container. It is also related to the \textit{Costs} entity where we store the distance and estimated duration of the trip between containers. This information is extracted with Google Maps Api v3 and is used to solve the underlying VRP. The generation of the duration and distance matrix is essential to generate a realistic solution. 

The table \textit{Forecasts} stores the predictions generated by the machine-learning algorithms that are used to estimate which containers will be collected in the next service. To undertake a service, the company has vehicles that have different sizes and loading capacities, which we must consider as constraints. A smaller truck has less loading capacity; thus it usually collects fewer containers.

Finally, once the execution of the route optimization algorithm has finished, we store the best solution in the database for visualization purposes. The solution of the whole problem is an ordered list of containers that must be visited by each vehicle. This information is stored in the entities \textit{Solution} and \textit{Tasks}. 


\subsection{Intelligent Decisions Module}

The fundamental characteristic of our software is the ability to predict the daily waste contribution for each of the containers that are part of the analyzed set. We have carried out an individualized mathematical modeling of each container using Weka's machine learning library~\citep{Frank2016}. Specifically we have used the time series analysis, which is the process of using statistical techniques to model and explain a time-dependent series of data points. Therefore, time-series forecasting is the process of using a model to generate predictions (forecasts) for future events based on known past events, like the fill-level of each container. The time-series forecasting algorithms enable the tool to generate filling predictions for those containers not monitored with sensors.

In this module we have also defined the inclusion/exclusion criteria used to make decisions about the containers that will be collected using the generated routes. The inclusion/excursion criteria can be defined according to container capacity, forecasts made, and real readings of the sensors, if possible. In addition, given a set of containers and a criterion, we have the possibility of forcing the inclusion of some containers with higher priority on the routes, and some may be excluded due to the defined criteria and constraints.

\subsection{Routes Generation Module}

This module has, as its main functionality, the generation of optimal routes for the waste collection. Given a set of containers, we use an evolutionary algorithm EA(1+1) based on the principle of ruin and recreation~\citep{Schrimpf2000} that allows us to generate efficient collection routes, taking into account the distance and the driving time between them.

In general, all the containers have the same capacity, but as time goes on, city managers buy new containers with potentially new, different capacities. Our software considers containers of different sizes and, therefore, different capacities. The capacity of each one is managed individually, and this allows us to estimate the total waste that the vehicle will collect, since we cannot exceed the load limit. In addition, vehicle fleets are not only made up of identical vehicles, but vehicle fleets are heterogeneous, so vehicles are also considered individually. For example, we can assign a larger number of containers to vehicles with higher capacity.

Sometimes there are special cases where the compatibility between container and vehicle has to be taken into account. This is the case when a container has to be picked up by a small vehicle. Our software allows us to take this restriction into account in order to generate valid routes. In addition, this feature can also influence the container collection time, which is the time it takes to the driver to pick up the container once it is in front of it. This input time is assigned depending on the location or characteristics of the container to be collected. This characteristic has a beneficial effect on the fact that the total estimated collection time is more realistic.

\subsubsection{Algorithm}

The evolutionary algorithm EA(1+1) used to solve the routing problem is based on the ruin and recreate principle. It is a large neighborhood search that combines elements of simulated annealing and threshold-accepting algorithms. This approach is suitable for complex problems that have many constraints~\citep{Schrimpf2000,Misevicius2003}. It is an all-purpose metaheuristic that can be used to solve a number of classical VRP types and basic search strategies can be easily varied to small and large movements according to the complexity of the problem. The pseudocode of the algorithm is shown in Algorithm~\ref{alg}.

\begin{algorithm}[h]
\caption{Pseudocode of the algorithm.}

\raggedright \textbf{Inputs:} Set of containers (location, fill-level, predictions), set of vehicles, distance matrix, duration matrix, deposit location \\
\textbf{Output:} Set of ordered lists of containers \\
\hrule

\begin{algorithmic}[1]
\label{alg}

\STATE bestSolution $\leftarrow$ randomSolution()
   
\WHILE{ $evals$ $<$ $totalEvals$} 
\STATE initialSolution $\leftarrow$ randomSolution()
\STATE removedTasks$\leftarrow$ \textbf{ruin}(initialSolution)
\STATE newSolution$\leftarrow$ reCreation(removedTask)
\STATE \textbf{evaluate}(newSolution)
\IF {(newSolution<bestSolution)}
\STATE bestSolution$\leftarrow$newSolution
\ENDIF
\ENDWHILE
\RETURN bestSolution
\end{algorithmic}
\end{algorithm}

The algorithm starts with a random initial solution. It divides parts of the solution leading to a set of tasks (collecting a container) that are not going to be assigned to a vehicle and a partial solution containing all other tasks. This process is called ruin the solution because we extract some tasks from the solution. Based on the partial solution, the algorithm reintroduces the tasks extracted leading to a new solution. Hence, this is the recreation process. If the new solution has more quality than the best-so-far solution, it is accepted as the new best solution. These steps are repeated until the algorithm reaches a number of iterations.

\subsection{Visualization Module}

 \begin{figure*}[]
        \centering
        \includegraphics[width=14cm]{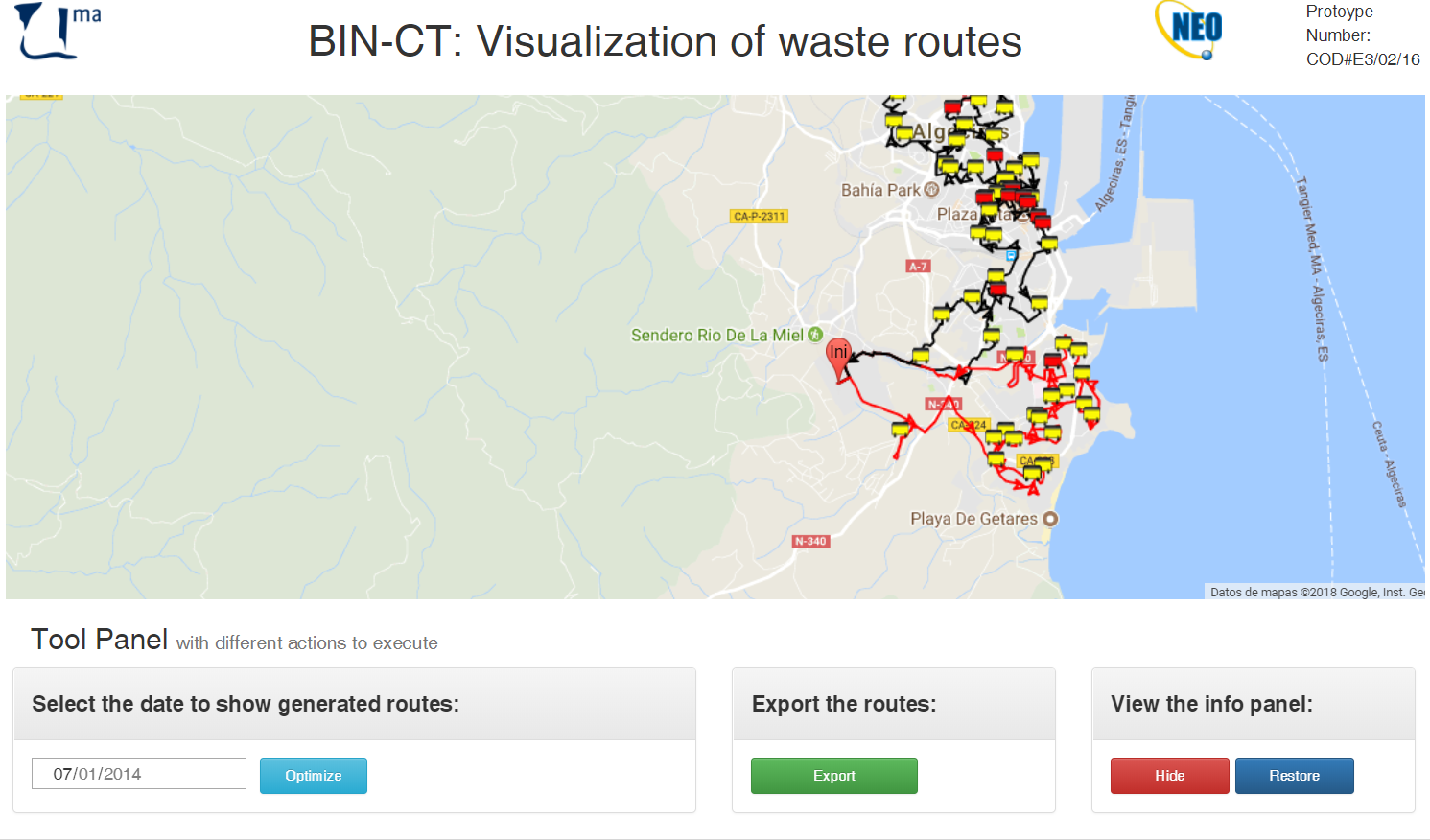}
        \caption{BIN-CT App interface.}
        \label{fig:app}
    \end{figure*}

The solutions are shown in an HTML navigable map. The software generates solutions for one or more days, and these can be viewed in a web browser. In Figure~\ref{fig:app} we show the interface of the web application, which can also be viewed online at \textit{http://mallba3.lcc.uma.es/binct/}. The main part shows a solution obtained for two vehicles: one red route and another black route, with two different trucks. In addition, we can observe the containers marked in red, which will be those whose prediction indicates a percentage of filling higher than 80\%.

The solutions can be accessed from any mobile device and the user interface adapts itself to the platform on the fly. This feature is especially desirable when the trucks' drivers have any doubts regarding the route they should follow.

\section{Experiments}
\label{experiments}

In this paper we carry out two different experiments. First, we focus on the predictions, which is the most innovative part of our system. They will allow us to achieve the greatest savings in the operation of our waste collection service. This first experiment consists in the comparison of three different algorithms to generate predictions. 

The second experiment is the generation of routes for a working day, starting from the real state that the containers had, so that, they have different filling states. These fill levels are complemented by the predictions for the next day. We compare our solution with the real solution provided by the company's human experts for that exact day (what really happened that day). In this way, we try to show that our approach is realistic, and therefore our solutions are too. As quality measures we analyze the distance, duration, and amount of waste collected using the generated routes. As a performance measure we report the time required to run the algorithm. We completed 30 independent runs in order to provide a fair value for the runtime. Finally, the experiments were run on a machine with Intel Core i7-3770 processor at 3.40~GHz and 16~GB memory.

\subsection{Case of Study}
\label{casestudy}

In this article we illustrate the behavior of BIN-CT with a real case study of an Andalusian city (Spain), where we highlight the benefits of our approach, showing it to also be effective and realistic. Our case study considers 217 paper containers from the metropolitan area of the city. The choice of an instance of recycling waste is more attractive than an organic waste collection. In this sense, we show the quality of our approach because most paper containers do not need to be emptied every day like organic waste ones, so they have a high variability in collection frequency. The problem instance parameters used in the case study are shown in Table~\ref{tab:case-study}.

\begin{table}[h]
\centering
\caption{Case study main parameters.}
\label{tab:case-study}
\begin{tabular}{lr}
\hline
Parameter          & Value      \\ \hline
Regular containers               & 208 units  \\
Small truck containers & 9 units    \\
Total containers & 217 units \\
Container capacity & 75 kg \\
Trucks                 & 2 units    \\
Unload container         & 3.5 min    \\
Small truck capacity       & 1700 kg \\
Big truck capacity       & 2000 kg \\
Historical data          & 11 months \\ \hline
\end{tabular}
\end{table}

\subsection{Experiment 1: Comparison of Time Series Algorithms}
\label{experiment1}

This experiment is intended to carry out a comparison between three time-series algorithms used to forecast the fill-level for all containers. The algorithms used in this comparison are: Linear Regression~\citep{Su2012}, Gaussian processes~\citep{Mackay1998} and Support Vector Machine for regression (SMOreg)~\citep{Rivas-Perea2013}. We briefly describe the techniques on which the algorithms are based:
\begin{itemize}
\item Linear regression works by estimating coefficients for a line that best fits the training data. It is a good idea to evaluate linear regression on a problem as a baseline, before moving onto more complex algorithms. 
\item Gaussian process is a collection of random variables, any finite number of which have a joint Gaussian distribution. A Gaussian process is completely specified by a mean function and a positive definite covariance function.
\item SMOreg works by finding a line of best fit that minimizes the error of a cost function. This is done using an optimization process that only considers those data instances in the training dataset that are closest to the line with the minimum cost.
\end{itemize}

In this experiment we use the fill-level of all 217 containers in 11 months. The data is obtained by the truck driver when the container is unloaded. If the container is not visited, then there is no datum for that specific day. Thus, the system must deduce what happened on those days for which we do not have data.

In order to alleviate this issue, we have performed a pre-processing of the data that has allowed us to obtain the estimated daily load increase of the containers. From the processed data, we have generated the daily predictions for the following month using the three forecasting algorithms.

The results indicate that the learning algorithm based on Gaussian processes is the best algorithm in our comparison, obtaining an average of just 3.83\% mean absolute error in the filling predictions of next month. Regarding the other techniques, the algorithm based on linear regression has an average of 7.41\% mean absolute error and the SMOreg algorithm has 9.52\%. After performing the Kruskal-Wallis statistical test (with Bonferroni correction to compare more than two samples) with 95\% confidence, we can state that there are significant differences between the results of the algorithm based on Gaussian processes with the other two algorithms of the comparison.

As the main conclusion of the first experiment, we can infer that the estimation provided by the best algorithm is more than acceptable, with less than 4\% mean absolute error. Note that a container is scheduled for the next collection service when the filling estimation is greater than 80\%, so the error is so negligible as to not affect the decision to collect a container. After analyzing these results, we decided to use the algorithm based on Gaussian processes for the second experiment.

\subsection{Experiment 2: The Daily Use of the System}
\label{experiment2}

\begin{table*}[h!]
\centering
\caption{Comparison between the real solution and the solution generated by our system. We highlight the best result for each indicator.}
\label{tab:results}
\begin{tabular}{llrrrr}
\hline
Solution                  & Truck   & Containers & Duration (s) & Distance (m) & Collected (kg) \\ \hline
\multirow{4}{*}{Experts}  & Small   & 31         & \textbf{12,044.00}        & 34,554.00          & 1,665.00       \\
                          & Big     & 31         & \textbf{13,912.00}        & 37,799.00          & 1,838.00       \\ \cline{2-6} 
                          & Total   & 62         & \textbf{25,956.00}        & 72,353.00          & 3,503.00       \\  
                          & Average &     -       & 418.64       & 1,166.98        & \textbf{56.50}      \\ \hline
\multirow{4}{*}{BIN-CT} & Small   & \textbf{33}         & 13,191.00         & \textbf{32,407.00}          & \textbf{1,689.00}       \\
                          & Big     & \textbf{41}         & 13,960.00        & \textbf{25,281.00}          & \textbf{1,984.00 }      \\  \cline{2-6} 
                          & Total   & \textbf{74}         & 27,151.00        & \textbf{57,688.00}          & \textbf{3,673.00}       \\ 
                          & Average &       -     & \textbf{366.90}       & \textbf{779.57}         & 49.63  \\ \hline  
\end{tabular}
\end{table*}

In the second experiment, we want to illustrate the general behavior of the system considering all the parameters specified in Table~\ref{tab:case-study}. In this experiment we generate routes for the first working day of the twelfth month starting from the real state of the containers and then we compare our solution with the real solution implemented that day.
Once our system has given the predictions for the following day, we obtain the estimated fill-level for all containers. Next, BIN-CT chooses the subset of containers that fulfills the following inclusion/exclusion criteria. Those containers with a fill-level higher than 50 \% are considered optional, while those with a filling greater than 80 \% are marked for mandatory collection.The application of this criterion to the set of 217 containers results in a subset of 77 containers to be picked up by two vehicles (the smallest is able to pick up the nine containers that cannot be collected by the big truck).

In Table~\ref{tab:results} we show the containers collected, the duration and distance of the routes, and the load of the trucks in the experts solution and the solution provided by BIN-CT. In addition, we show the totals of the measures for both vehicles and the average per container. After the execution of the route generation algorithm, we obtain two different routes: the biggest truck should pick up 41 containers and the smallest truck should visit 33 containers. Therefore, our solution is able to schedule most of the containers with only 3 containers not scheduled for the next service due to truck capacity constraints.

The system provides very accurate information on the duration of the routes, the duration of the four routes can be seen in the fourth column in Table~\ref{tab:results}. On the one hand, the route of the small truck has an estimated duration of 3 hours, 39 minutes and 51 seconds to collect a total of 33 containers. On the other hand, the route of the big truck will last 3 hours, 52 minutes and 40 seconds to collect a total of 41 containers. However, if we take a look at the distance, the small truck's route is 32.4 km compared with just 25.2 km of the other route. Therefore, it seems that the constraint of picking up those 9 containers, which cannot be collected by the big truck, is a detrimental to the small truck's route. Regarding the trucks load, the amount of waste collected is proportional to each truck's size.

Let us compare the experts solution with our solution. First of all, we analyze how many containers, which are also in the expert’s solution, were selected by BIN-CT. The result is that 66\% of the experts’ solution's containers are also collected in the solution provided by BIN-CT. This means that using static plans, sometimes they collect containers that are below 50\% according to our estimations. Actually, around 20\% of the containers (12 units) of the studied day had less than 50\% fill-level. Our system is therefore unique in detecting them and saving costs in the planning.

The routes provided by BIN-CT are able to pick up more containers, in less time per container, traveling fewer kilometers and taking more advantage of the load capacity of the trucks. We must highlight the difference in distance traveled. We would save 20\% of the distance traveled in comparison with the real solution even though our routes took 19 minutes and 55 seconds more than the real ones. If we used the average distance to collect a container and multiply that by the 62 containers of the real solution, then the estimated distance traveled would be 48,333 meters, which means we would have 33.2\% shorter routes than the routes used by the company. The algorithm runs with the default parameters and a stop criterion of 10,000 iterations. The median runtime was 27,350 milliseconds that is less than half a minute, so the algorithm is quite fast. Moreover, if we increase the iteration budget, potentially we would obtain even better results.

After specialized staff from the company, who gave us the data for the case of study, validated the solutions generated by BIN-CT, we can conclude that our system generates realistic solutions and reproduces with enough reliability, what happens in reality. In addition, the use of the fill predictions discarded 140 containers not included as nodes to visit in the generation of routes process. Actually, all containers (except one) which were not included in our routes but included in the real ones, were discarded based on the forecasting and not because of truck capacity constraints. This fact means that with high probability we have saved unnecessary visits to semi-empty containers that are part of a predefined static route for the day analyzed. Therefore using BIN-CT implies a reduction in both the cost of the routes and the harmful emissions thrown by the trucks.
	
\section{Conclusions and Future Work}
\label{conclusions}

Traditional solid waste management is usually based on previous experience rather than data. Currently the waste collection plans are static with a predetermined number of visits per week and designed manually, which implies a slow adaptation to any change in normal conditions. 
In this paper we have proposed BIN-CT, an intelligent predictive software system designed to use knowledge extracted from historical data and technology that is part of the state-of-the-art in urban sustainability research. We are convinced that our intelligent system will help to improve the quality of life of the city where it is implemented. The reduction of costs and the efficient treatment of the waste generated will constitute a way to achieve sustainability in the city.

Our aim is to reduce the costs of the waste collection service, while increasing the quality of service for the citizen. Our software suggests the collection of containers above 80 \% and thus avoids the annoying overflows of containers. Through the study case that we have analyzed in this article, we prove that the solutions generated are realistic. At the same time, they are more efficient because the solutions prevent trips to semi-empty containers and the generated routes are 33.2\% shorter than those used by the company. Moreover, after the comparison of three different algorithms, we used the best which generates predictions with just a 3.83\% mean absolute error. This fact indicates that the forecasting performed is fairly reliable for the calculations we need to make in our system.

BIN-CT has a direct application for waste management companies, both in Spanish cities and elsewhere in the world. This non-commercial software system, based on computational learning algorithms, does not require a large investment in infrastructure, which is why it is highly attractive to companies. There is also an interesting scientific side (use of time series, predictions, efficient algorithms, …) that companies may be interested in, which could make the leap in quality and distinction that is sought after in a market as competitive as the current one.

As for future work, we highlight the possible integration of an indeterminate number of sensors, in a larger cyber-physical system, to measure the amount of waste in the containers. The way the system is built facilitates an easy integration of sensors. The use of sensors makes it possible to change a route in real time by adding or removing containers. Fortunately, this valuable feature could save more unnecessary trips to containers, while at the same time preventing possible overflows of containers.

Regarding the algorithms for the generation of predictions, here we have studied some techniques for time-series prediction based on regression. However, recently the recurring neural network technique has emerged as a good method for the generation of time-series predictions~\citep{LeCun2015,Camero2018}. We plan to improve the fill-level predictions by using an algorithm based on recurring neural networks.


\section*{Acknowledgments}
This research has been partially funded by the Spanish Ministry of Science and Innovation and FEDER under contracts RTC-2017-6714-5 and TIN2017-88213-R and the network of smart cities CI-RTI (TIN2016-81766-REDT). The CELTIC project C2017/2-2 under contracts \#8.06/5.47.4997 y \#8.06/5.47.4996. J. Ferrer thanks University of M\'alaga for his postdoc fellowship. 

\section*{References}

\bibliographystyle{elsarticle-harv.bst}

\bibliography{sample.bib}

\begin{thebibliography}{34}
\expandafter\ifx\csname natexlab\endcsname\relax\def\natexlab#1{#1}\fi
\expandafter\ifx\csname url\endcsname\relax
  \def\url#1{\texttt{#1}}\fi
\expandafter\ifx\csname urlprefix\endcsname\relax\def\urlprefix{URL }\fi

\bibitem[{Al-Salem et~al.(2009)Al-Salem, Lettieri, and Baeyens}]{Al-Salem2009}
Al-Salem, S.~M., Lettieri, P., Baeyens, J., oct 2009. {Recycling and recovery
  routes of plastic solid waste (PSW): A review}. Waste Management 29~(10),
  2625--2643.

\bibitem[{Arribas et~al.(2010)Arribas, Blazquez, and Lamas}]{Arribas2010}
Arribas, C.~A., Blazquez, C.~A., Lamas, A., apr 2010. {Urban solid waste
  collection system using mathematical modelling and tools of geographic
  information systems}. Waste Management {\&} Research 28~(4), 355--363.

\bibitem[{Beli{\"{e}}n et~al.(2014)Beli{\"{e}}n, {De Boeck}, and {Van
  Ackere}}]{Belien2014}
Beli{\"{e}}n, J., {De Boeck}, L., {Van Ackere}, J., feb 2014. {Municipal Solid
  Waste Collection and Management Problems: A Literature Review}.
  Transportation Science 48~(1), 78--102.

\bibitem[{Buenrostro-Delgado et~al.(2015)Buenrostro-Delgado, Ortega-Rodriguez,
  Clemitshaw, Gonz{\'{a}}lez-Razo, and
  Hern{\'{a}}ndez-Paniagua}]{Buenrostro-Delgado2015}
Buenrostro-Delgado, O., Ortega-Rodriguez, J.~M., Clemitshaw, K.~C.,
  Gonz{\'{a}}lez-Razo, C., Hern{\'{a}}ndez-Paniagua, I.~Y., 2015. {Use of
  genetic algorithms to improve the solid waste collection service in an urban
  area.} Waste management (New York, N.Y.) 41, 20--7.

\bibitem[{Camero et~al.(2018)Camero, Toutouh, Stolfi, and Alba}]{Camero2018}
Camero, A., Toutouh, J., Stolfi, D.~H., Alba, E., 2018. {Evolutionary Deep
  Learning for Car Park Occupancy Prediction in Smart Cities}. In: Learning and
  Intelligent OptimizatioN Conference LION.

\bibitem[{Cole et~al.(2014)Cole, Quddus, Wheatley, Osmani, and Kay}]{Cole2014}
Cole, C., Quddus, M., Wheatley, A., Osmani, M., Kay, K., feb 2014. {The impact
  of Local Authorities' interventions on household waste collection: a case
  study approach using time series modelling.} Waste management (New York,
  N.Y.) 34~(2), 266--72.

\bibitem[{Dantzig and Ramser(1959)}]{Dantzig1959}
Dantzig, G.~B., Ramser, J.~H., 1959. {The Truck Dispatching Problem}.
  Management Science 6~(1), 80--91.

\bibitem[{Dat et~al.(2012)Dat, {Truc Linh}, Chou, and Yu}]{Dat2012}
Dat, L.~Q., {Truc Linh}, D.~T., Chou, S.~Y., Yu, V.~F., 2012. {Optimizing
  reverse logistic costs for recycling end-of-life electrical and electronic
  products}. Expert Systems with Applications 39~(7), 6380--6387.

\bibitem[{Dijkstra(1959)}]{Dijkstra1959}
Dijkstra, E.~W., 1959. {A note on two problems in connexion with graphs}.
  Numerische Mathematik 1~(1), 269--271.

\bibitem[{Frank et~al.(2016)Frank, Hall, and Witten}]{Frank2016}
Frank, E., Hall, M.~A., Witten, I.~H., 2016. {The WEKA Workbench}.

\bibitem[{Garvin et~al.(2011)Garvin, Cohen, and Dwyer}]{GCD11}
Garvin, B.~J., Cohen, M., Dwyer, M.~B., 2011. {Evaluating improvements to a
  meta-heuristic search for constrained interaction testing}. Empirical
  Software Engineering 16~(1), 61--102.

\bibitem[{Ghisellini et~al.(2016)Ghisellini, Cialani, and
  Ulgiati}]{Ghisellini2016}
Ghisellini, P., Cialani, C., Ulgiati, S., 2016. {A review on circular economy:
  The expected transition to a balanced interplay of environmental and economic
  systems}. Journal of Cleaner Production 114, 11--32.

\bibitem[{Gruler et~al.(2017{\natexlab{a}})Gruler, Fikar, Juan, Hirsch, and
  Contreras-Bolton}]{Gruler2017}
Gruler, A., Fikar, C., Juan, A.~A., Hirsch, P., Contreras-Bolton, C., feb
  2017{\natexlab{a}}. {Supporting multi-depot and stochastic waste collection
  management in clustered urban areas via simulation–optimization}. Journal
  of Simulation 11~(1), 11--19.

\bibitem[{Gruler et~al.(2017{\natexlab{b}})Gruler, Fikar, Juan, Hirsch, and
  Contreras-Bolton}]{Gruler2017-2}
Gruler, A., Fikar, C., Juan, A.~A., Hirsch, P., Contreras-Bolton, C., feb
  2017{\natexlab{b}}. {Supporting multi-depot and stochastic waste collection
  management in clustered urban areas via simulation–optimization}. Journal
  of Simulation 11~(1), 11--19.

\bibitem[{Gruler et~al.(2017{\natexlab{c}})Gruler, Fikar, Juan, Hirsch, and
  Contreras-Bolton}]{Gruler2017-3}
Gruler, A., Fikar, C., Juan, A.~A., Hirsch, P., Contreras-Bolton, C., feb
  2017{\natexlab{c}}. {Supporting multi-depot and stochastic waste collection
  management in clustered urban areas via simulation–optimization}. Journal
  of Simulation 11~(1), 11--19.

\bibitem[{Hansmann and Zimmermann(2009)}]{Baudach}
Hansmann, R.~S., Zimmermann, U.~T., 2009. {Integrated Vehicle Routing and Crew
  Scheduling ( IVRCS ) in Waste Management Part I}. Dagstuhl Seminar
  Proceedings 09261 Models and Algorithms for Optimization in Logistics, 1--8.

\bibitem[{LeCun et~al.(2015)LeCun, Bengio, and Hinton}]{LeCun2015}
LeCun, Y.~A., Bengio, Y., Hinton, G.~E., 2015. {Deep learning}. Nature
  521~(7553), 436--444.

\bibitem[{Mackay(1998)}]{Mackay1998}
Mackay, D. J.~C., 1998. {Introduction to Gaussian processes}. Neural Networks
  and Machine Learning 168~(1996), 133--165.

\bibitem[{Misevicius(2003)}]{Misevicius2003}
Misevicius, A., 2003. {Genetic algorithm hybridized with ruin and recreate
  procedure: Application to the quadratic assignment problem}. In:
  Knowledge-Based Systems. Vol.~16. pp. 261--268.

\bibitem[{Nuortio et~al.(2006)Nuortio, Kyt{\"{o}}joki, Niska, and
  Br{\"{a}}ysy}]{Nuortio2006}
Nuortio, T., Kyt{\"{o}}joki, J., Niska, H., Br{\"{a}}ysy, O., 2006. {Improved
  route planning and scheduling of waste collection and transport}. Expert
  Systems with Applications 30~(2), 223--232.

\bibitem[{Ombuki-Berman et~al.(2007)Ombuki-Berman, Runka, and
  Hanshar}]{Ombuki-Berman2007}
Ombuki-Berman, B.~M., Runka, A., Hanshar, F.~T., 2007. {Waste collection
  vehicle routing problem with time windows using multi-objective genetic
  algorithms}. Proceedings of the Third IASTED International Conference on
  Computational Intelligence, 91--97.

\bibitem[{Rabbani et~al.(2018)Rabbani, Heidari, Farrokhi-Asl, and
  Rahimi}]{Rabbani2018}
Rabbani, M., Heidari, R., Farrokhi-Asl, H., Rahimi, N., jan 2018. {Using
  metaheuristic algorithms to solve a multi-objective industrial hazardous
  waste location-routing problem considering incompatible waste types}. Journal
  of Cleaner Production 170, 227--241.

\bibitem[{Rivas-Perea(2013)}]{Rivas-Perea2013}
Rivas-Perea, P., 2013. {Support Vector Machines for Regression: A Succinct
  Review of Large-Scale and Linear Programming Formulations}. International
  Journal of Intelligence Science 03~(01), 5--14.

\bibitem[{Sakti et~al.(2018)Sakti, Yu, and Sopha}]{Sakti2018}
Sakti, S., Yu, V.~F., Sopha, B.~M., 2018. {Heterogeneous Fleet Location Routing
  Problem for Waste Management: A Case Study of Yogyakarta, Indonesia}. In:
  International Conference in Management Science and Decision Making. pp.
  1--11.
\newline\urlprefix\url{https://repository.ugm.ac.id/274607/1/2018{\_}ICSDM.pdf}

\bibitem[{Schrimpf et~al.(2000)Schrimpf, Schneider, Stamm-Wilbrandt, and
  Dueck}]{Schrimpf2000}
Schrimpf, G., Schneider, J., Stamm-Wilbrandt, H., Dueck, G., 2000. {Record
  Breaking Optimization Results Using the Ruin and Recreate Principle}. Journal
  of Computational Physics 159~(2), 139--171.

\bibitem[{Sniezek and Bodin(2006)}]{Sniezek2006}
Sniezek, J., Bodin, L., apr 2006. {Using mixed integer programming for solving
  the capacitated arc routing problem with vehicle/site dependencies with an
  application to the routing of residential sanitation collection vehicles}.
  Annals of Operations Research 144~(1), 33--58.

\bibitem[{Su et~al.(2012)Su, Yan, and Tsai}]{Su2012}
Su, X., Yan, X., Tsai, C.-L., 2012. {Linear regression}. Wiley
  Interdisciplinary Reviews: Computational Statistics 4~(3), 275--294.

\bibitem[{Tavares et~al.(2009)Tavares, Zsigraiova, Semiao, and
  Carvalho}]{Tavares2009}
Tavares, G., Zsigraiova, Z., Semiao, V., Carvalho, M., mar 2009. {Optimisation
  of MSW collection routes for minimum fuel consumption using 3D GIS
  modelling}. Waste Management 29~(3), 1176--1185.

\bibitem[{Teixeira et~al.(2004)Teixeira, Antunes, and de~Sousa}]{Teixeira2004}
Teixeira, J., Antunes, A.~P., de~Sousa, J.~P., nov 2004. {Recyclable waste
  collection planning––a case study}. European Journal of Operational
  Research 158~(3), 543--554.

\bibitem[{Tukker(2015)}]{Tukker2015}
Tukker, A., 2015. {Product services for a resource-efficient and circular
  economy - A review}. Journal of Cleaner Production 97, 76--91.

\bibitem[{Tung and Pinnoi(2000)}]{Tung2000}
Tung, D.~V., Pinnoi, A., 2000. {Vehicle routing-scheduling for waste collection
  in Hanoi}. European Journal of Operational Research 125~(3), 449--468.

\bibitem[{Ustundag and Cevikcan(2008)}]{Ustundag2008}
Ustundag, A., Cevikcan, E., 2008. {Vehicle Route Optimization for Rfid
  Integrated Waste Collection System}. International Journal of Information
  Technology {\&} Decision Making 07~(04), 611--625.

\bibitem[{Varone et~al.(2014)Varone, Markov, and Bierlaire}]{Varone2014}
Varone, S., Markov, I., Bierlaire, M., 2014. {Vehicle Routing for a Complex
  Waste Collection Problem Iliya Markov, Ecole Polytechnique
  F{\'{e}}d{\'{e}}rale de Lausanne Vehicle Routing for a Complex Waste
  Collection Problem}. Tech. rep.

\bibitem[{Xu et~al.(2013)Xu, Gao, Cui, and Liu}]{Xu2013}
Xu, L., Gao, P., Cui, S., Liu, C., jun 2013. {A hybrid procedure for MSW
  generation forecasting at multiple time scales in Xiamen City, China.} Waste
  management (New York, N.Y.) 33~(6), 1324--31.

\end{thebibliography}

\end{document}